# An Enhanced Latent Semantic Analysis Approach for Arabic Document Summarization

Kamal Al-Sabahi[1] · Zuping Zhang[1] · Jun Long[1] · Khaled Alwesabi[1]

[1]School of Information Science and Engineering, Central South University, Changsha, China

k.alsabahi@csu.edu.cn; zpzhang@csu.edu.cn

**ABSTRACT**

The fast-growing amount of information on the Internet makes the research in automatic document summarization very urgent. It is an effective solution for information overload. Many approaches have been proposed based on different strategies, such as latent semantic analysis (LSA). However, LSA, when applied to document summarization, has some limitations which diminish its performance. In this work, we try to overcome these limitations by applying statistic and linear algebraic approaches combined with syntactic and semantic processing of text. First, the part of speech tagger is utilized to reduce the dimension of LSA. Then, the weight of the term in four adjacent sentences is added to the weighting schemes while calculating the input matrix to take into account the word order and the syntactic relations. In addition, a new LSA-based sentence selection algorithm is proposed, in which the term description is combined with sentence description for each topic which in turn makes the generated summary more informative and diverse. To ensure the effectiveness of the proposed LSA-based sentence selection algorithm, extensive experiment on Arabic and English are done. Four datasets are used to evaluate the new model, Linguistic Data Consortium (LDC) Arabic Newswire-a corpus, Essex Arabic Summaries Corpus (EASC), DUC2002, and Multilingual MSS 2015 dataset. Experimental results on the four datasets show the effectiveness of the proposed model on Arabic and English datasets. It performs comprehensively better compared to the state-of-the-art methods.

**KEYWORDS**

Latent semantic analysis, Automatic text summarization, Singular value decomposition, Adjacent Weight, Part of Speech ((POS

## 1 Introduction

Information are playing an important role in everyone's daily life. The rapid development of information technology improves the quantity and the complexity of information sources, especially that related to text documents [1] (e.g., scientific papers, blogs, books, portal news, opinion extraction, etc.) [2]. In this massive amount of information, it is very difficult to choose the relevant information that meets the needs of the user [3]. The search engine is required to perform a significant task to provide the user with a subset of the original data. Unfortunately, this subset is still substantial in size. The user still needs to go through each single item until he/she finds the information of his interest. This boring task makes automatic document summarization very urgent. The summary allows the user to decide whether a document is relevant to his/her interest, rather than reading the entire document [4]. In addition, other systems, such as news portals, search engines, intelligent gathering systems, question answering, can use summarization techniques to improve their efficiency and performance. The key point behind all these techniques is to find a representative subset of the original document such that the core essence of the document is contained in this subset from the semantic and conceptual standpoints [5]. Achieving this is very challenging because it is difficult to determine the quality of the summary, since it depends on several factors such as the user's requirements, the user's background, and the compression ratio among others. However, there is some success to tackle these problems where the scientists have been able to reach a level in which the machine is able to generate a human readable summary, especially for English documents. Unfortunately, Arabic document summarization is still receiving a little attention. Due to the lack of

Arabic open-source tools and resources (e.g., thesaurus, corpora, part of speech taggers, morphological analyzers, online dictionaries, etc.), there is a huge gap between the research work in English and Arabic [7]. Despite the promising results of English summarizers, there is still much work to be done before automatic summarizers catch up with human beings [6].

Recently, significant research efforts have been devoted to solve the documents summarization problem. Many approaches have been proposed ranging from statistical based algorithms to graph-based and machine learning algorithms [8]. Many of these systems address the summarization problem depending on the required application and the user's needs [7]. There are two common types of automatic text summarization, extractive and abstractive. Extractive text summarization techniques perform text summarization by selecting the important sentences from the original text and combine them into a new shorter version according to some criteria [9]. The sentence selection process is usually based on linguistic, mathematical and statistical techniques. The extractive summary may not be coherent, but it delivers the main idea about the input text [10]. In contrast, abstractive text summarization techniques attempt to build an internal semantic representation of the original text and then create a summary closer a human-generated one. The state-of-the-art abstractive models are still quite weak, so most of the previous work has focused on the extractive summarization [11]. A fundamental requirement in any extractive summarization model is to identify the salient sentences that represent the key information mentioned in the document [12]. In text summarization problem, large feature sets are a challenge that should be handled for better performance. Therefore, utilizing feature reduction techniques are important for efficient representation of textual features. This concern has sparked a great interest in using latent semantic analysis (LSA) to solve the summarization problem.

LSA is a powerful unsupervised analytical tool. It is one of the most prominent learning algorithm in Information Retrieval tasks. It has the ability to reveal the unseen structure of words, among words, sentences or text through singular value decomposition (SVD). It also produces measures of word–word, word–document and document–document relations that are well correlated with several human cognitive phenomena involving association or semantic similarity. The performance of LSA-based summarization algorithms depends on the quality of the document representation [13] and the sentence selection algorithm. Earlier LSA-based document summarization approaches have some limitations. In this work, we try to overcome those limitations by improving the document's representation and proposing a new sentence selection algorithm.

The main contribution of this work is a novel LSA-based approach for Arabic text summarization that can capture the latent semantic structure of a document to generate a coherent summary with a good diversity and coverage. The key difference to previous work is that our model addresses several issues that are not adequately modeled by the previously proposed models. First, LSA-based algorithms do not use information of word order, morphologies and syntactic relations. These kinds of information are necessary for finding out the semantic meaning of the text, so in the proposed approach, the adjacent weight is considered while building the input representation matrix which in turn helps to resolve the sparsity of the data. Second, LSA-based approaches suffer from the high dimensionality and vulnerability to noise; a POS tagger is used to tackle these issues. Finally, to ensure the informative and the diversity of the generated summary, a new selection algorithm is proposed. The new algorithm combines term description with sentence description for each topic to ensures that the candidate sentences are the best representation of the topic. The new selection algorithm differs from the previous ones in the fact that the three yield matrices of SVD ($U$, $\Sigma$, $V$) are involved in the selection process. To our knowledge, we are the first one to use the three matrices for this purpose. Experimental results show that our model outperforms previous approaches by a significant margin for Arabic and English on the four datasets used in this work.

The rest of this paper is organized as follows: the related work is presented in Sect. 2. In Sect. 3, we describe in detail the proposed approach including the annotation details and the generation of the summary. Experiments, test collection, evaluation metrics and experimental results are presented in Sect. 4. Finally, we discuss the results and conclude in Sect. 5.

## 2 Related Work

Document summarization is one of the most difficult, though promising, application of natural language processing (NLP). The researchers have been striving to utilize any advancement in NLP to create a more efficient summary. Extractive summarization techniques are the most common ones, whose basic idea is to extract important sentences from text documents and then recombine them to form a summary [14]. Recently, several summarization models have been proposed. Some of them are based on LSA.

LSA-based document summarization approaches usually go through three main stages, Input Matrix Creation, SVD, and Sentence Selection. Almost all these approaches perform the first two phases of LSA algorithm in the same way. There is a bit difference in the weighting schemes used to fill out the input matrix. Another difference is in the way they select sentences for the summary, sentence selection algorithm. In this section, we briefly review the work higher relevant to the study of this paper, including: LSA selection algorithms and the previous Arabic LSA-based approaches.

## 2.1 LSA Sentence Selection Algorithms

After building the input matrix $A$ and calculating SVD, three matrices are yield, $U$, $\Sigma$ and $V^T$. In $V^T$ matrix, the columns represent the sentences and the rows represent the concepts. The row order shows the importance of the concepts, so the first row in $V^T$ determines the most important one. The matrix cell values indicate how strong the relationship between sentences and concepts. Gong and Liu [15] have used the $V^T$ to select the candidate sentences. They just simply chose one sentence for each row of the matrix. For example, the first chosen sentence is the one with the largest index value in the first row of $V^T$, and the second sentence is the one with the largest index value in the second row of $V^T$ and so on. Gong and Liu's [15] method has some drawbacks. First, it assumed that the chosen concepts are in the same level of importance. If the summary compression rate is high, sentences from less important concepts are chosen. In contrast, important concepts might have sentences that are highly related to them, but they do not have the highest cell value, so choosing only one sentence for each concept might be not enough to represent that concept.

Steinberger and Jezek [16] tried to overcome the disadvantages of Gong and Liu algorithm [15]. In their approach, they let more that one sentence to be chosen for each important concept. For this purpose, they used both $V^T$ and $\Sigma$ matrixes for sentence selection. Only concepts whose indexes are less than or equal to a predefined number are considered. They used the values of the singular matrix, $\Sigma$, as a multiplication parameter to give more importance to the concepts that are highly related to the text. This model lacks a method of deciding how many LSA dimensions to include in the latent space, where they assumed that the dimension of the new space is given as a parameter to the algorithm by the user.

Murray et al. [17] follow the Gong and Liu approach [15], but they let more that one sentences to be chosen for each topic rather than selecting one. The number of the selected sentences per topic is determined by the percentage of the related singular value over the sum of all singular values in $\Sigma$ matrix. The number of the selected sentences for each topic is learned from the data to some extent. One of the drawback of this method is that if there is a big difference between the current singular value and the next one, the topics whose singular values are less than the current one may not be included.

Ozsoy et al. [18] proposed a new approach for sentence selection called a cross-method. Their approach is an extension to the approach of Steinberger and Jezek [16]. They used $V^T$ matrix for sentence selection purpose. A pre-processing step exists between the SVD and the sentence selection. The claimed aim of the pre-processing step is to remove the overall effect of the sentences that are related to the concept somehow, but not one of the most significant sentences for that concept. After pre-processing, they fellow Steinberger and Jezek approach with some modification.

Another LSA-based summarization algorithm was proposed by Wang and Ma [19]. In which, they used the neighbor weight of the term in two neighbor sentences to the current sentence while building the input matrix. After applying SVD, they used the two matrices $U$ and $V^T$ to select sentences for the summary. Some of the shortage of this approach is the fact that it selects three sentences at most from each topic. This fixed number may be unreasonable especially for long documents, where an important topic may need more than three sentences to have a fair representation in the summary and the opposite for the less important topics, as we will discuss in Section 3.4.1.c.

Another LSA-based study was carried out by Shen et al. [20]. In which, they tried to capture the important contextual information for latent semantic modeling by proposing a latent semantic model based on the convolutional neural network with convolution-pooling structure, called the convolutional latent semantic model (CLSM). The CLSM first presents each word within its context to a low-dimensional continuous feature vector, which directly captures the contextual features at the word n-gram level. Then, the CLSM discovers and aggregates only the salient semantic concepts to form a sentence level feature vector. After that, the sentence level feature vector is further fed to a regular feed-forward neural network, which performs a nonlinear transformation, to extract high-level semantic information of the word sequence. However, this method was not basically proposed to solve document summarization problem.

As we discussed earlier in this section, the previous models have some disadvantages. The major weakness of the current LSA-based summarization approaches is the using of weighting schemes that mostly failed to produce a good document represenation. In addition, there is a limitation of their sentence selection algorithms, as we will discuss further in Section 3.4.1.c.

## 2.2 Arabic LSA Models

The research on automatic English text summarization has started since more than 50 years ago, while studies of the Arabic language in this field began appearing only in the last decade [21]. Several approaches have been proposed ranging from surface level approaches to machine learning ones. However, most of these studies followed the numerical approach. In this section, we focus on the ones that are the most related to our work.

El-Haj et al [22] proposed a multi-document summarization model for Arabic and English. They used LSA to measure the similarity between the extracted sentences to reduce the redundancy in the generated summary. The impact of Arabic text summarization on Arabic document clustering has been investigated by Froud et al. [23]. They used LSA to generate Arabic summaries, then they used those summaries to represent the document in Vector Space Model to enhance the Arabic document clustering. The experimental results of this study are related to text clustering, not summarization. Their experimental results show that text summarization improved the clustering process. However, in terms of purity values, unpromising results were obtained when applying the summarization process. Another previous studies was carried out by Ba-Alwi et al. [24]. In which, they investigate the impact of the Arabic word variations and the different weighting techniques on the performance of the Arabic LSA-based text summarization. While selecting the sentences for the summary, they used the same algorithm that was proposed by Wang and Ma [19]. One of the raised questions about the findings of the paper is that the ROUGE-2 scores are unbelievably high. In some cases, they are near to the ROUGE-1 scores. They did not give much details about the implementation and the evaluation settings. All the previous models used the existing sentence selection algorithms that discuss earlier in Sect. 2.1. This means that they inherit most of the drawbacks of those algorithms.

### 2.3 Arabic Natural Language Processing Challenges

Arabic has some similarity with other languages such as English; however, it also has its own characteristics that make Arabic NLP more sophisticated and challenging. There are many aspects that justify the slow progress in Arabic NLP tools, especially when it comes to text summarization [25]. First, Arabic complex morphology, both derivational and inflectional, makes it hard to do a morphological analysis [8]; since it may consist of prefixes, lemma and suffixes in different combinations. Second, word senses disambiguation in Arabic is very challenging. A word can have up to seven synonyms, each of which meaning depends on the context [26]. Third, borrowing is commonly used in Arabic, where a certain word can be borrowed to be used in a different context. This makes it difficult for a machine to sense the meaning. Another challenge is the fact that Arabic does not have orthographic representation of short vowels. In Arabic text, the diacritics are usually omitted, which makes it difficult to correctly tokenize and parse the text [27]. The absence of capitalization and the minimal punctuation is another issue that makes the identification of proper nouns, titles, abbreviations, and acronyms very hard. On the top of that, there is also a shortage of Arabic lexicons, corpora, machine-readable dictionaries, and Arabic gold standard summaries [21].

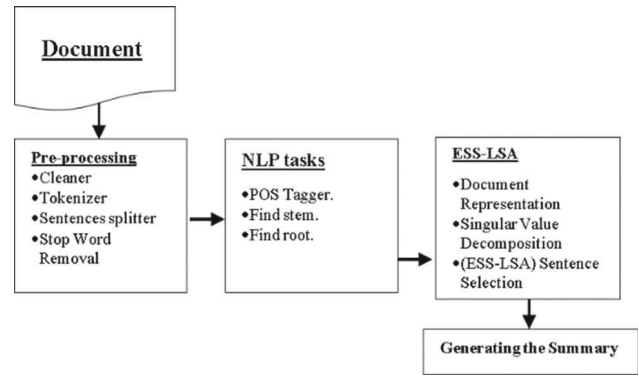

Fig. 1 The proposed approach phases

## 3 LSA-Based Approach for Document Summarization

In this work, a new LSA-based model is proposed. The enhanced LSA-based text summarization framework provides summaries with reasonable quality compared to the existing models. The model consists of four main steps: Pre-processing, NLP tasks (stemming, finding the root and POS tagging), LSA implementation, and then generating the summary. The next section explains these stages in more details.

### 3.1 The Proposed Approach Phases

The document is preprocessed first, then analyzed and synthesized before the result, which is the summary, is generated. External tools have been used to aid in several tasks during the pre-processing and the analyzing stages, namely stem, root, POS tagger. Figure 1 shows the major stages and their subcomponents as implemented in the proposed model.

#### 3.1.1 Pre-processing

The summarization process requires many pre-processing tasks before generating the summary. The most important one is removing stop words. Stop words refer to the most common words in a language which do not contribute meaning to determine the important content of the document. In this paper, a predefined list of stop words that contains 1377 words [28] is used.

#### 3.1.2 Tokenizer and Sentences Splitter

Each document is decomposed into individual sentences; this decomposition is a source of ambiguity because punctuation is rarely used in Arabic texts and when it exists, it is not always critical to guide the decomposition. In addition, some words can mark the beginning of a new sentence (or proposition). For this purpose, two kinds of decomposition are used.

| The Arabic Morpho syntactic Markers and Functional Words | ادوات الربط والوصل في اللغة العربية |
|---|---|
| In, and, then, or, but, when | في،و ،ثم،أو ،أم،بل،لكن،لكن،حتى |
| Also, after, although, as before, but this, not | ايضا، بعد،بالرغم،حيث،قبل،ولهذا ، وليس |

**Fig. 2** Samples of Arabic morpho-syntactic markers and functional words [23]

A morphological decomposition based on punctuation and decomposition based on the recognition of markers morpho-syntactic or functional words such as: أو, و , لكن،حتى, or, and, but, when. However, these particles may play a role other than to separate phrases as shown in Fig. 2.

In this work, Arabic sentences split using punctuation marks that define the end of each sentence. A set of punctuation marks, including commas (,) semicolons (;), question marks (?), exclamation marks (!), colons (:), and periods (.), are selected to split the text into sentences. Words are split using the space. Sentences and word boundaries are detected with the help of python packages from the source sentences.

### 3.1.3 Natural Language Processing Tasks

As mentioned earlier, some NLP tasks have been applied over the corpus, parts of which are pre-processing. In the following sections, another two important NLP tools are used to enhance the output of the proposed model and investigate the impact of each of them in the final document representation. The motivation behind this investigation is the fact that most of the previous studies in Arabic NLP applications used the stemming like what is usually done in English. However, in Arabic, the stem and the root of a word are not always the same, so may be using stem while pre-processing Arabic text is not enough.

**(1) Arabic word morphology**

In Arabic, word derivation is represented in three concepts: root, pattern and word form. The word forms, such as verbs, verbal nouns, agent nouns, etc., can be obtained from roots by applying derivational rules. Generally speaking, each pattern carries a meaning combined with the meaning inherited from the root, which in turn gives the target meaning of the lexical form [27]. According to the level of analysis, Arabic stemming algorithms can be categorized into root-based approach and stem-based approach. Root-based approach uses morphological analysis to extract the root of a given Arabic word; while Stem-based approach removes the most frequent suffixes and prefixes. Different approaches for Arabic stemming can be identified, such as manually constructed dictionaries, algorithmic light stemmer, and morphological analyzers. Al-Fedaghi and Al-Anzi [23] algorithms try to find the root of the word by matching the word with all possible patterns and affixes attached to it. In this paper, ISRI Arabic Stemmer in python NLTK is used to stem Arabic words. The algorithm for this stemmer is described in [29].

**(2) POS Tagger**

POS tagger is the process of identifying the parts of speech for each word in a text [30]. In this work, POS tagger proposed by Köprü [31] is applied over the input text. Tokenized words are annotated with POS tags that are used in subsequent stages. Where AD is an adjective, NN is a noun, AV is an adverb and VB is a verb. Only terms with NN* and VB* tags (nouns and verbs from any kind) are considered on the experiments that used POS.

### 3.1.4 Latent Semantic Analysis (LSA) for Document Summarization

Usually, there are three main steps for any LSA-based algorithm: creating the input matrix (sentence–term matrix), applying SVD to matrix and selecting the sentences for the summary.

**(a) Input Matrix Creation**

The input document is represented as an $m \times n$ matrix ($A$). Each row in $A$ matrix represents a term and each column represents a sentence $A = [a_{1j}, a_{2j}, \ldots a_{nj}]$. The cell value ($a_{ij}$) represents the importance of the word. In this work, the entry $a_{ij}$ is obtained by multiplying a local weight $L(t_{ij})$ and global weight $G(t_{ij})$, then adding adjacent weight ($W_{adj}$). Equations (1) and (2) are used to calculate the weight of every cell in the input matrix ($A$).

$$a_{ij} = L(t_{ij}) \times G(t_{ij}) + W_{2\text{Adj}(t_{ij})} \quad (1)$$

$$a_{ij} = L(t_{ij}) \times G(t_{ij}) + W_{4\text{Adj}(t_{ij})} \quad (2)$$

where $L(t_{ij})$ is the Local Weight for the $i$th term in the $j$th sentence, $G(t_{ij})$ is the Global Weight for the $i$th term in the whole document, and $W_{2\text{Adj}(tij))}$, $W_{4\text{Adj}(tij))}$ are the adjacent weights for the $i$th term in a window of two and four adjacent sentences to the $j$th sentence, respectively.

There are different weighting schemes to calculate the local weight, the global weight, and the adjacent weight for the $i$th term in the $j$th sentence. These approaches are as follows:

- **Local Weight**

There are four alternative strategies for calculating the local weight [32], as follows:

(a) *Binary Representation (BR):* If term ($t_i$) exists in sentence ($j$), $L(t_{ij}) = 1$, otherwise $L(t_{ij}) = 0$.
(b) *Term Frequency (TF):* $L(t_{ij}) = tf_{ij}$, where $tf_{ij}$ refers to the number of times that $t_i$ term appears in sentence ($j$).
(c) *Augment weight (AW):* The augment weight is calculated by Eq. (3):

$$L(t_{ij}) = 0.5 + 0.5 \times \left(\frac{tf_{ij}}{tf_{\max}}\right) \quad (3)$$

Where $tf_{ij}$ denotes the number of times that the $i$th term occurs in the $j$th sentence and $tf_{\max}$ denotes to the frequency of the most frequent term in the $j$th sentence.

(d) *Logarithm Weight (LW):* The logarithm weight is calculated by Eq. (4):

$$L(t_{ij}) = 1 + \log(tf_{ij}) \quad (4)$$

- **Global Weight**

Global Weight can be calculated using one of the following alternatives [19]:

(a) *No Global Weight (NG):* $G(t_{ij}) = 1$.
(b) *Inverse Sentence Frequency (ISF):*

$$G(t_{ij}) = 1 + \log\left(\frac{n}{n_i}\right) \quad (5)$$

where $n_i$ is the number of sentences that contain the term ($t_i$) and $n$ is the total number of sentences.

(c) *Entropy Frequency (EF)*

The Entropy Frequency is calculated by Eqs. (6) and (7):

$$G(t_{ij}) = 1 + \sum \frac{P_{ij} \log P_{ij}}{\log n}, \quad (6)$$

where

$$P_{ij} = \frac{tf_{ij}}{gf_i} \quad (7)$$

where $tf_{ij}$ refers to the number of times that the term ($t_i$) occurs in the $j$th sentence, $n$ denotes the total number of sentences, and $gf_i$ refers to the number of times that the term ($t_i$) appears in the entire document.

- **Adjacent Weight**

The weight of the term ($t_i$) in the adjacent sentences is considered during the calculating of the value of $a_{ij}$ for $t_i$ in the $j$th sentence. In this work, two kinds of adjacent weight are considered, near and far (Definition 1 & Definition 2), respectively. Before proceeding further, let us introduce some notations and definitions.

**Definition 1** For the input document (D) with n sentences and m terms, let D= $(S_1, S_2, \ldots, S_n)$ and $T_j = (t_{1j}, t_{2j}, \ldots, t_{ij})$ where $t_i$ ($1 \leq i \leq m$) refers to the $i$th term, and $S_j$ ($1 \leq j \leq n$) refers to the $j$th sentence. Let $L(t_{i,j-1})$ and $G(t_{i,j-1})$ are the local weight and the global weight, respectively, for the term ($t_i$) in the left adjacent sentence. $L(t_{i,j+1})$ and $G(t_{i,j+1})$ are the local weight and the global weight, respectively, for the term ($t_i$) in the right adjacent sentence. For each $t_i$ ($1 \leq i \leq m$), the two-adjacent weight for the term $t_i$ in $S_j$ is computed by Eq. (8).

$$\begin{aligned} W_{2\text{Adj}(t_{ij})} = \gamma \big[ &L(t_{i,j-1}) * G(t_{i,j-1}) \\ &+ L(t_{i,j+1}) * G(t_{i,j+1}) \big] \end{aligned} \quad (8)$$

where $\gamma$ is a parameter with a range of value between 0 and 1.

**Definition 2** Let $L(t_{i,j-2}), G(t_{i,j-2}), L(t_{i,j-1})$ and $G(t_{i,j-1})$ are the local weight and the global weight, respectively, for the term ($t_i$) in two left adjacent sentences. $L(t_{i,j+2}), G(t_{i,j+2}), L(t_{i,j+1})$ and $G(t_{i,j+1})$ are the local and global weight, respectively, for the term ($t_i$) in two right adjacent sentences. For each $t_i$ ($1 \leq i \leq m$), the four-adjacent weight for the term $t_i$ in $S_j$ is calculated by Eq. (9)

$$\begin{aligned} W_{4\text{Adj}(t_{ij})} = \gamma \big[ &0.5 \times L(t_{i,j-2}) \times G(t_{i,j-2}) + L(t_{i,j-1}) \\ &\times G(t_{i,j-1}) + L(t_{i,j+1}) \times G(t_{i,j+1}) + 0.5 \\ &\times L(t_{i,j+2}) \times G(t_{i,j+2}) \big] \end{aligned} \quad (9)$$

*Remark 1* By experiment, we found that the important of the two adjacent weights are not the same, so we set the far adjacent weight ($j \pm 2$) to be equal to the half of the near ($j \pm 1$), as shown in Eq. (9):

*Remark 2* By experiment, the best value for $\gamma$ is 0.5.

Adjacent Weight is added mainly by the following three notable considerations:

(1) There is a relationship between adjacent sentences which makes the topics more convincing.
(2) A pronoun and what it refers to usually appear in the adjacent sentences, so considering these relationship helps to resolve anaphora resolution.
(3) Adjacent weight scheme helps to resolve the issue of data sparsity.

**(b) Singular Value Decomposition**

Singular Value Decomposition SVD is an algebraic method that can identify the relationships between words and sentences [33].

**Theorem 1** *Let A is an m × n matrix in K. Then there is a factorization, called a singular value decomposition of A, of the form* [34]*:*

$$A = U\Sigma V^T \qquad (10)$$

Where

- $K$ is either the field of real numbers or the field of complex numbers.
- $U$ is an $m \times n$, unitary matrix,
- $\Sigma$ is a diagonal $n \times n$ matrix with nonnegative real numbers on the diagonal, and
- $V$ is a $n \times n$, unitary matrix over $K$. $V^T$ is the conjugate transpose of $V$.

Once the term by sentence matrix ($A$) is constructed. We used *scipy.sparse.linalg* python library to apply SVD on the input matrix $A$.

$U$, $\Sigma$ and $V^T$ are the three matrices mentioned in Theorem 1. $U$ is an $m \times n$ matrix which represents term by concept, $\Sigma$ is an $n \times n$ diagonal matrix with nonnegative real numbers on the diagonal which represent the scaling values. The magnitude of singular values in $\Sigma$ suggests the degree of importance of the concepts. $V^T$ is an $n \times n$ real or complex unitary matrix which represents concept-by-sentence as shown in Eq. (10).

**(c) The Proposed Algorithm for Sentence Selection**

The output of SVD is used to select the candidate sentences for the summary. There are many algorithms to determine the summary sentences. These algorithms are explained in Sect. 2.1. In this work, a new algorithm for sentence selection is proposed. It is a refinement of the algorithm presented by Wang and Ma [19]. We called the algorithm "An enhanced LSA sentence selection algorithm," shortly (ESS-LSA). In which many procedures are improved including the preprocessing of the input matrix $A$. In addition, the three matrices ($U$, $\Sigma$, and $V$) are utilized to select the candidate sentences based on the following considerations:

(1) In Wang and Ma [19] approach, there is a fixed number for the selected sentences for each topic, three sentences at most from each topic. In a very long document, the important topic may need to be represented by more than three sentences. If we take too few, we may lose topics which are important from a summarization point of view. In contrast, if we take too many, we end up including less important topics.

(2) The proposed algorithm is based on the hypothesis that $\Sigma$ matrix suggests the degree of importance of the concepts. The matrix ($\Sigma$) are used to determine the maximum number of sentences to be selected for each topic (concept) by getting the percentage of the related singular value over the summation of all singular values. In the algorithm, we call this the sentence-topic threshold ($P_k$), Definition 3. Before calculating the threshold, the singular values that are less than half of the largest singular value are set to zero, if any exists.

(3) Using the root of the word is supposed to improve the performance of the algorithm by increasing the semantic similarity and reduce the dimension.

(4) The weights of the term in four adjacent sentences, two sentences from the right and two from the left to the current sentence, are added to the weighting schemes that build the input matrix $A$.

(5) The algorithm proposed by Wang and Ma [19] depends on the hypothesis that the most representative sentences of the current concept should include the terms that best represent this concept, Proposition 2. As a result, if there is any noise in the input matrix, it will lead to select unimportant sentences. Under this concern, POS tagger is applied during the creation of the input matrix. The terms with NN* and VB* tags (nouns and verbs from any kind) are considered.

**Definition 3** Let $E$ is the $n \times n$ diagonal matrix of SVD which suggests the degree of importance of the concepts. Then, let k is the number of concepts that can be selected and $N_k$ is the number of sentences for the $k$th topic. $P_k$ is the sentence-topic thresholds as in Eq. (11):

$$P_k = \frac{E[k,k]}{\Sigma_{i=1}^{n} E[i,i]} \times n \qquad (11)$$

Before calculating the threshold ($P_k$) as in Definition 3, the singular values in $E$ that are less than the minimum sigma value are set to zero. The aim of this is to remove the effect of the very low singular values. The minimum sigma value is calculated by Eq. (12):

$$\text{MinSigmaValue} = \max(E) * \alpha \qquad (12)$$

where α is a constant value. By experiment, the best value is 0.5 which mean that the minimum sigma value used in this work is equal the half of the largest singular value.

*Remark 3* To avoid confusion between the summation and singular value matrix $\Sigma$ in Definition 3 and Algorithm 1, $E$ is used instead of $\Sigma$ to denote the singular values matrix.

**Proposition 1** *A concept is represented by a few terms, and these terms must have the largest index values in*

*the corresponding left singular vector. The two forms of description of a concept are called concept-by-sentence and term-by-concept description. Each concept is treated as an independent topic.*

**Proposition 2** *The most representative sentences of the current concept should include the terms that best represent this concept. Therefore, each topic in the summary can be rebuilt by selecting sentences according to the important of the index values in the right singular vector until a few of most representative terms that have the largest index values in the left singular vector are fully included with $P_k$ threshold of sentences from each topic.*

The process of selecting sentences for the summary is explained in the following two subsections and the complete algorithm is shown in Algorithm 1.

- *Formulation.* Let $D$ is a document with $n$ sentences, and $m$ terms. The maximum number of the candidate sentences is $M$; let $k$ is the number of concepts to be selected, while $N_k$ is the number of sentences for the $k$th concept. The initial values for $k$ and $N_k$ are 1 and 0, respectively. Assume $S$ contains the sentences of the summary and initialized to null.
- *Sentence Selection and Term Selection.*

The sentence selection algorithm contains the following steps:

(1) Getting the sentence-topic threshold $P_k$ by calculating the percentage of the related singular value over the summation of all singular values in the normalized $E$ matrix, Definition 3.
(2) While the number of sentences in the summary is less than $M$, for the $k$th concept, we select the sentence that has the largest index value from the $k$th right singular vector $v_k$.
(3) Then, we get $c$ that satisfies $v_{kc} = \text{Max}(v_{ki})$, add the $c$th sentence $sent_c$ into $S$ and remove the $c$th element $v_{ic}$ for $v_i$ ($c \leq i \leq n$). Next, update $V^T$ and increase $N_k$.
(4) Then, we select the top three largest index values from the three terms ($u_{kp}, u_{kq}, u_{kr}$) from the $k$th left singular vector $U_k$, and set $T = \text{term}_p, \text{term}_q, \text{term}_r$. Then, remove terms that appear in both $sent_c$ and $T$ from $T$.
(5) While $T$ is not null, $N_k$ is less than the sentence-topic threshold $P_k$ and the number of sentences in the summary is less than $M$, go to steps 2. otherwise set $T$ to null and increase $k$ and start selecting sentences for the next concept.

## 3.2 Algorithm Description and Complexity Analysis

The aim of the proposed algorithm is to select candidate sentences for the summary using the three matrices generated by SVD. The complexity depends on the execution time of the innermost loop in Algorithm 1 (line 17–25). That loop basically selects sentences for the summary. It is executed every time we need to add a new candidate sentence to the summary set. Taking a closer look at Algorithm 1 reveals that line 2–5 are only used for calculating the maximum number of the selected sentences for each topic by getting the percentage of the related singular value over the summation of all singular values. The first step is to find the maximum singular value in $E$ matrix. Second, this value is used to normalize the singular matrix by setting all the singular values less that the half of the maximum value to zero. Then the summation of all singular values is calculated. The time complexity for this is $O(n)$. The execution of the innermost loop depends not only on $P_k$ but also on whether there are still elements in $T$ or not. We will limit our investigation to the worst case only, so the time complexity for this loop is $O(P_k)$. Adding a sentence into the summary happens only once for each sentence, while the number of the selected sentences never exceeds $M$. As we see, selecting the sentence happens once, whether in the outer loop or in the inner loop. That means the value of $|S|$ is increasing for each repetition of the two loops. Therefore, the complexity depends on the execution time of the basic operation which is the selection of the candidate sentences for the summary. Selecting $M$ sentences for the summary takes at most $O(n) + O(P_k) + O(M)$, where $O(n)$ is the execution time for the first loop in the algorithm, lines (2–5), $O(P_k)$ is the execution time of the inner loop, lines (17–28) and $O(M)$ is the execution time for the main loop, lines (8–31); hence, the overall time complexity of the algorithm is estimated as $O(n)$.

Taking a close look at the previous algorithms, that we have mention in Sect. 2.1, [15–19], we find that they share the same basic operation despite their difference on the amount of calculations in each algorithm, but all of which have a complexity bound to $O(n)$. It is the same complexity that our proposed algorithm has, but in our algorithm, the input size is much smaller. The pre-processing steps, the POS tagger, and using the root dramatically reduce the dimension of the input matrix and consequently the output of SVD. That enhances the performance of the proposed algorithms and decreases the time for processing, especially for a huge input.

*Algorithm 1. An enhanced LSA Sentence Selection Algorithm (ESS-LSA)*

**Input**: Document, $U$, $E$ & $V^T$, Number of document sentences ($n$), Number of summary sentences ($M$)

**Output**: the generated summary ($S$).

**Procedure** SentenceSelection

| | |
|---|---|
| 1 | Set $S$ =null,  $k$: =1, $N_k$: =0, $Sum_E$: =0 |
| 2 | *MinSigmaValue* := max($E$) * 0.5 |
| 3 | $E[E < MinSigmaValue] = 0$ |
| 5 | $Sum_E$ :=tr($E$)   // finding the sum of the elements on the main diagonal |
| 8 | **While** $|S| < M$ |
| 9 |     Get the index of Max value ($c$) in $V_k$ row |
| 10 |     $P_k := (E[k,k]/sum_E) * n$    // Calculate Sentence-Topic Threshold $P_k$ from $E$ matrix |
| 11 |     Delete $V_c$ from $V^T$ |
| 12 |     Set $N_k := N_k + 1$ |
| 13 |     Get Max ($p,q,r$) in $U_k$, $T$:=$term_p, term_q, term_r$ |
| 14 |     $T_0 := T \cap Sent_c$, $T := T - T_0$ |
| 15 |     $S$: =$S$ + $Sent_c$ // add the sentence to the summary |
| 16 |     Remove $Sent_c$ from *DocumentSentences* list |
| 17 |     **While** $T \neq \emptyset$ **do** |
| 19 |         If ($N_k \leq P_k$ & $|S| < M$) |
| 20 |             Get the index of the Next Max value ($c$) in $V_k$ row. |
| 21 |             $S$: =$S$ + $Sent_c$ // add the sentence to the summary |
| 22 |             Remove $Sent_c$ from *DocumentSentences* list |
| 23 |             Remove $V_c$ from $V^T$ |
| 24 |             Set $N_k := N_k + 1$ |
| 25 |             $T_0 := T \cap Sent_c$, $T := T - T_0$ |
| 26 |         **Else** |
| 27 |             Set $T: = \emptyset$ |
| 28 |         **End if** |
| 29 |     **End while** |
| 30 |     Set $k$: =$k+1$ |
| 31 | **End while** |
| 32 | Return $S$ |

## 4 Experiments and Results

For comparison, four datasets are used in this work for Arabic and English. The datasets are EASC, Arabic LDC, the well-known DUC2002 Single-Document Summarization dataset, and from Multilingual 2015 dataset, the experiment is carried out on Single-document Summarization (MSS) for English and Arabic. The focus of this study is proposing a new model for Arabic document summarization. The aim of using English datasets is to evaluate the proposed LSA-based selection algorithm in another language other than Arabic. The goal is to answer the following questions:

(1) What is the optimal weighting scheme for creating the LSA input matrix?
(2) What is the effect of using word morphology namely root, stem, and word?
- Word: using all words-excluding stop words- as it is in the input without changes.

- Stem: by removes the most common prefixes and suffixes from words.
- Root: uses pattern matching to extract the roots.

(3) What is the effect of using POS tagger on the performance of the proposed LSA approach? We will consider only NN* & VB* (nouns and verbs from any kind).
(4) What is the effect of considering the weights of the term in four adjacent sentences, two from the right and two from the left, on the performance?
(5) What is the results of the comparison between our model and the baselines on DUC2002 and MultiLing MSS 2015?

For each document, several combinations of models are used, each combination is a separate experiment as shown in Table 1:

Evaluating the generated summaries is one of the most difficult and costly tasks in text summarization. It is difficult to come to a judgement whether a summary has a good quality since there is no agreement about the evaluation criteria that should be adopted [35]. Previous studies have shown that, human summarizers tend to agree only about 60% of the times, and in only 82% of the cases humans agreed with their own judgment [36]. Summary evaluation approaches try to determine how much the summary is adequate, reliable, useful and relative to its source [37]. In this work, two methods are used to evaluate the generated summaries:

(1) Human judgment for each generated summary for Arabic LDC.
(2) ROUGE-1 and ROUGE-2.

All the experiments described in this work were evaluated using the ROUGE Toolkit[1]. ROUGE-1 and ROUGE-2 [38] are applied. On the four datasets, different latent semantic analysis models are applied using different input matrices.

*Remark 4* To ensure that the recall-only evaluation will be unbiased to length, the option "-l 100" is used in ROUGE to truncate longer summaries in DUC2002. For MSS 2015 dataset, output summaries are truncated to the size of the articles' human summaries.

### 4.1 Arabic Document Summarization

The Arabic test collection for the proposed approaches involve two datasets. The first one is delivered by the Linguistic Data Consortium (LDC) at the University City of PENN USA. Arabic LDC dataset includes two Arabic collections,

**Table 1** The experiments abbreviation and explanation (the abbreviations are explained in section

| The experiment | Explanation |
| --- | --- |
| RAWEF2ADJ | Root & (AW) * (EF) + ($W_{2Adj(tij)}$) |
| RAWEF4ADJ | Root & (AW) * (EF) + ($W_{4Adj(tij)}$) |
| RAWEFPOS2ADJ | Root & POS & (AW) * (EF) + ($W_{2Adj(tij)}$) |
| RAWEFPOS4ADJ | Root & POS & (AW) * (EF) + ($W_{4Adj(tij)}$) |
| RAWISF4ADJ | Root & (AW) * (ISF) + ($W_{4Adj(tij)}$) |
| RAWISFPOS4ADJ | Root & POS & (AW) * (ISF) + ($W_{4Adj(tij)}$) |
| RBREF2ADJ | Root & (BR) * (EF) + ($W_{2Adj(tij)}$) |
| RBREF4ADJ | Root & (BR) * (EF) + ($W_{4Adj(tij)}$) |
| RBREFPOS2ADJ | Root & POS & (BR) * (EF) + ($W_{2Adj(tij)}$) |
| RBREFPOS4ADJ | Root & POS & (BR) * (EF) + ($W_{4Adj(tij)}$) |
| RLWEF4ADJ | Root & (LW) * (EF) + ($W_{4Adj(tij)}$) |
| RLWEFPOS4ADJ | Root & POS & (LW) * (EF) + ($W_{4Adj(tij)}$) |
| SAWEF4ADJ | Stem & (AW) * (EF) + ($W_{4Adj(tij)}$) |
| SAWEFPOS4ADJ | Stem & POS & (AW) * (EF) + ($W_{4Adj(tij)}$) |
| WAWEF4ADJ | Original Word & (AW) * (EF) + ($W_{4Adj(tij)}$) |

Arabic GIGAWORD and Arabic NEWSWIRE-a corpus. It contains 100 documents categorized into 5 reference sets; each contains 20 documents about the same topic. In this work, five documents are randomly chosen from each set at a total of 25 document. For each document, fifteen summaries are generated within a compression ratio of 30%. The second dataset is Essex Arabic Summaries Corpus (EASC) [39,40]. EASC includes 153 Arabic articles in 10 topics. For each document, five model extractive summaries are available. These model summaries were generated by native Arabic speakers using Mechani fcal Turk (http://www.mturk.com/)[2]. The compression rate for the generated summaries is 40% [40].

#### 4.1.1 Manual Evaluation for Arabic Document Summaries (Human Judgment)

Since there are golden/reference summaries, automatic techniques are used for evaluating summaries. However, we choose to evaluate the output summaries using human judgments to have another way of evaluation for the generated summaries despite the expensiveness and time-consuming of the manual evaluation. A total of 50 Arabic specialists from different backgrounds participated in the evaluation. Each participant was given a document with three summaries, one generated using root, the second using stem and the third using word. The summaries were handed to the participants in a random order. Each participant was asked to read the document and its summaries and then to evaluate each summary based on the TAC Responsiveness Metric consisting of

---

[1] ROUGE-1.5.5 with options -c 95 -m -n 2 -x.

[2] http://www.mturk.com/.

**Table 2** Evaluation Scale source interpretations

| Evaluation | Score | Interpretation |
|---|---|---|
| V. Poor | 0 | The summary is not related to the document at all |
| Poor | 1 | The core meaning of the document is missing |
| Fair | 2 | The user is somehow satisfied with the summary, but he/she expects more |
| Good | 3 | The summary is readable and it carries the main idea of the document |
| V. Good | 4 | The summary is very readable and focuses more on the core meaning of the document. The user is totally satisfied with the summary |

**Table 3** Human evaluation scores for the proposed approaches

| Method | Scores | | | | | Mean |
|---|---|---|---|---|---|---|
| | 0 | 1 | 2 | 3 | 4 | |
| | V. Poor | Poor | Fair | Good | V. Good | |
| Root | 0.00 | 0.00 | 0.25 | 0.50 | 0.25 | **3.00** |
| Stem | 0.25 | 0.25 | 0.33 | 0.17 | 0.00 | 1.42 |
| Word | 0.50 | 0.33 | 0.17 | 0.00 | 0.00 | 0.67 |

The bold value shows the highest scores, among the numbers in those tables

Content and Readability/Fluency measures [22]. The possible judgments, their scores, and the interpretations are given in Table 2.

The results obtained from the distributed questionnaires for the three kinds of models (Table 3, Fig. 3) show that the ranks and scores for the system that used root are comparatively better than the others two that used stem and word. The mean of scores are 3, 1.4 and 0.67 for root, stem and word, respectively. This was expected since using the root increases the semantic similarity between sentences.

### 4.1.2 Automatic Evaluation

The using of ROUGE as an automatic evaluation method on LDC Arabic Newswire-a dataset needs a reference or a model summary. As a part of the questionnaire, we asked more than 50 Arabic specialists to manually summarize the documents. Then those summaries are used as golden summaries. The respondents were asked to read and summarize a given document using the same words that exists in the original text by writing what they considered to be the most significant in the document. They were instructed to write no more than 30% of the original document. As a result, five summaries were created for each document in the collection. Each of the summaries for a given document were generated by five different Arabic specialists. The other three datasets have their own golden summaries.

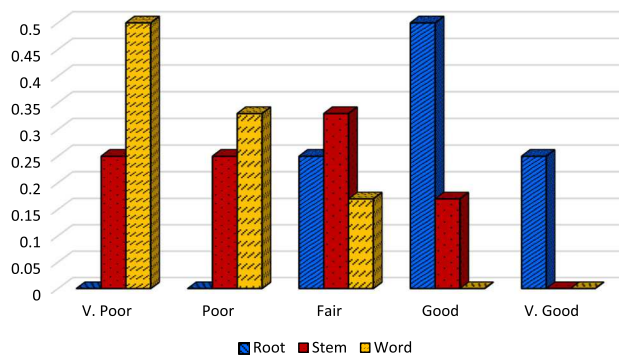

**Fig. 3** Human evaluation scores for the proposed approaches

### 4.1.3 Results and Discussion

To answer the experimental questions, raised in beginning of Sect. 4, ROUGE-1, ROUGE-2 are applied on the output of every experiment on LDC Arabic Newswire-a dataset with five different reference (Model), summaries written manually by humans. Moreover, the same experiments are implemented on the other dataset, EASC. The generated summaries are compared with the human-generated summaries in this dataset. As we discussed in Sect. 4 and Table 1, fifteen-combinations are used, each combination is a separate experiment.

From the obtained results, we can make the following observations that answer the experiment questions:

i. As shown in Table 4, it has been observed that the method that used root has the highest score among the other two, which used word and stem. Using the root of a word increases the semantic similarity between sentences and increases the value of Term Frequency ($tf$) through the document. This will consider all different styles of the word which have the same root as one term. This step increases the weight of the semantic feature in the sentence and gives better performance. This reveals one of the characteristics of Arabic where the stem and the root of a word are not always the same, so may be using the stem while pre-processing Arabic text is not enough.

ii. Figure 4 and Table 5 show that the best combination of weighting schemes is the combination Augment weight (AW) * Entropy Frequency (EF), it performs better than the other combinations on both datasets. The possible reason is that Augment weight and Entropy Frequency have more normalized formulas.

iii. Adding the adjacent weight to the weighting schemes has improved the performance of almost all the combinations. Figure 4 show that the using of $W_{4\text{Adj}(t_{ij})}$ in the weighting schemes for the input matrix outperforms the one using $W_{2\text{Adj}(t_{ij})}$. Considering the weights of the term in four adjacent sentences ($W_{4\text{Adj}(t_{ij})}$) played

**Table 4** The effect of word morphology on the Performance of the proposed method on Arabic LDC

| Model | Rouge-1 | Rouge-2 |
|---|---|---|
| ROOT(RAWEFPOS4ADJ) | **0.6175** | **0.3336** |
| STEM(SAWEFPOS4ADJ) | 0.5028 | 0.2811 |
| WORD(WAWEFPOS4ADJ) | 0.3564 | 0.1482 |

The bold values show the highest scores, among the numbers in those tables

**Table 5** The performance comparison of all models on Arabic LDC

| Model | ROUGE-1 | ROUGE-2 |
|---|---|---|
| RAWEF4ADJ | 0.4142 | 0.2107 |
| RAWEFPOS2ADJ | 0.5620 | 0.2771 |
| RAWEFPOS4ADJ | **0.6175** | **0.3336** |
| RAWISF4ADJ | 0.4160 | 0.2313 |
| RAWISFPOS4ADJ | 0.3861 | 0.2102 |
| RBREF2ADJ | 0.1749 | 0.0244 |
| RBREF4ADJ | 0.3993 | 0.2069 |
| RBREFPOS2ADJ | 0.3031 | 0.1737 |
| RBREFPOS4ADJ | 0.4039 | 0.2101 |
| RLWEF4ADJ | 0.3031 | 0.1251 |
| RLWEFPOS4ADJ | 0.3876 | 0.2041 |
| SAWEF4ADJ | 0.3727 | 0.1415 |
| SAWEFPOS4ADJ | 0.5028 | 0.2811 |
| WAWEFPOS4ADJ | 0.3564 | 0.1482 |

The bold values show the highest scores, among the numbers in those tables an important role in the performance of all the proposed models, since the using of $W_{4\text{Adj}(t_{ij})}$ resolves the issue of data sparsity and utilizes the syntactic relations.

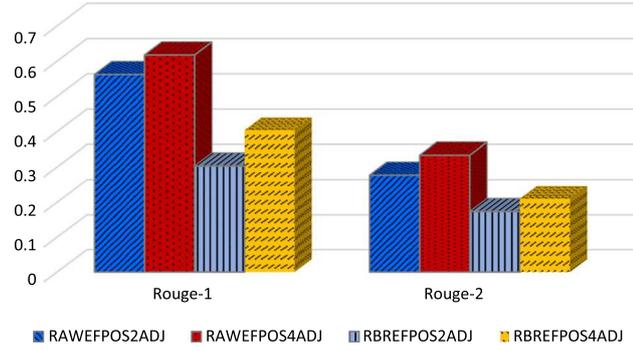

**Fig. 4** The effect of the four-adjacent weight on the performance compared to two-adjacent weight on Arabic LDC

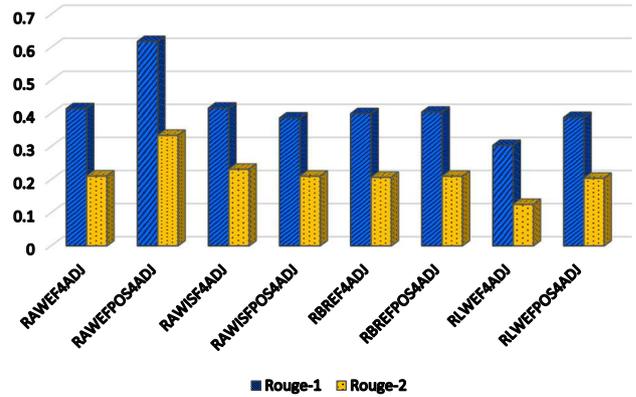

**Fig. 5** The effect of using POS on the performance of models on Arabic LDC

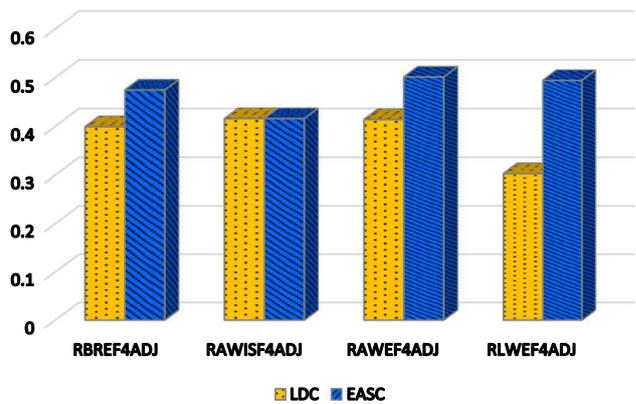

**Fig. 6** The performance comparison of different weighting schemes on the two datasets in the terms of ROUGE-1

iv. With POS tagging, almost all the models acquire an improvement. The using of POS tagger has enhanced the performance of LSA, where it achieved better results than the models that do not used POS as shown in Fig. 5. Applying POS tagger to the input text decreases the noise and tackles the high dimensionality of LSA.

v. Furthermore, the results of all the experiments conducted in this work, Table 5, assert that the model that uses the root of the word with POS tagging, Augment weight (AW) * Entropy Frequency (EF) + Adjacent weight ($W_{4\text{Adj}(t_{ij})}$) as a weighting scheme is the best model. It largely outperforms all other models applied to the Arabic test collection.

vi. Finally, Fig. 6 shows that the performance of the proposed models on the two datasets, Arabic LDC and EASC, leads to the same conclusion.

### 4.2 English Document Summarization

#### 4.2.1 Dataset

The proposed LSA-based algorithm also applied to English documents on the well-known dataset, DUC2002. The main goal is to evaluate the performance of the proposed LSA-Based sentence selection algorithm on other languages other than Arabic. In this experiment, the POS tagger for English is not used. DUC2002 dataset contains 567 news articles made available during the DUC 2002 evaluations, and the corre-

**Table 6** The ROUGE scores for English on DUC2002

| Model | ROUGE-1 | ROUGE-2 |
| --- | --- | --- |
| LEAD-3 | 0.436 | 0.210 |
| DUC-best | 0.498 | 0.252 |
| Steinberger2004 [16] | 0.433 | 0.180 |
| Wang2013 [19] | 0.472 | **0.261** |
| Ours(RAWEF4ADJ) | **0.500** | 0.241 |

The bold values show the highest scores, among the numbers in those tables sponding 100-word summaries for each document (single-document summarization), or the 100-word summaries for each of the 59 document clusters formed on the same dataset (multi-document summarization). In this work, the single-document summarization tasks are used. ROUGE-1 and ROUGE-2[3] are adopted for evaluation in this experiment.

#### 4.2.2 Baselines

For comparison, several baselines are used on DUC 2002. First, the leading sentences (Lead-3) is used as a baseline. It simply produces the leading three sentences of the document as the summary. Second, the BEST scores on DUC2002 is reported. Third, the proposed approach is compared with the approach proposed by Wang and Ma [19]. In Addition, the popular LSA-Based Summarization approach proposed by Steinberger and Jezek [16] is implemented and used as a baseline on this dataset for English

#### 4.2.3 Results and Discussion

As shown in Table 6, the proposed model outperforms the baselines on DUC2002 in the term of ROUGE-1, but it couldn't beat DUC-best and [19] in the term of ROUGE-2. The possible reason is that ROUGE-2 tends to score grammatically rather than contents [41]. In addition, the authors of the paper [19] didn't give details about the ROUGE toolkit options that they used to evaluate their model. The good ROUGE-1 result returns to the fact that the proposed method combines the term description with the sentence description for each topic which enhances the informativeness and the diversity of the generated summary.

### 4.3 Arabic/English Document Summarization

#### 4.3.1 Dataset

To check the effectiveness of the proposed method on Arabic and English languages, Multilingual 2015 Single-document

**Table 7** ROUGE scores for Arabic and English on MultiLing2015 dataset

| Model | English | | Arabic | |
| --- | --- | --- | --- | --- |
| | ROUGE -1 | ROUGE -2 | ROUGE -1 | ROUGE -2 |
| Worst | 0.3717 | 0.0993 | 0.5266 | 0.1765 |
| Best | 0.5038 | 0.1412 | 0.5741 | 0.2064 |
| Ours(RAWEF4ADJ) | **0.5159** | **0.1441** | **0.5759** | **0.2258** |

The bold values show the highest scores, among the numbers in those tables

Summarization (MSS) [42] task is used. MSS 2015[4] task was to generate single-document summaries for some selected Wikipedia articles with at least one out of 38 languages defined by organizers of the task. For each one of the 38 languages, there are 30 documents. In this work, the proposed method is evaluated using two languages, Arabic and English. ROUGE-1 and ROUGE-2[5] are adopted for evaluation in this experiment.

#### 4.3.2 Baselines

The baselines used for comparison on Multiling MSS 2015 are the BEST and The WORST scores obtained by the 23 participating systems on this dataset for Arabic and English.

#### 4.3.3 Results and Discussion

From the results in Table 7, it is worth mentioning the following notes. First, it can be observed that our proposed approach (RAWEF4ADJ) achieves better performance on MSS 2015 with respect to the worst and the best scores on this dataset, except ROUGE-2 on English. The good results assert the effectiveness of using the adjacent weight in the weighting schemes and adopting the sentence-topic threshold which enables the selection of variable number of sentences for each topic based on singular values in $\Sigma$ matrix.

## 5 Conclusion and Future Work

In this paper, an in-depth case analysis is performed on the capacity of the proposed models, through which the effect of the word morphology, weighting schemes, POS tagger, the adjacent weight and the sentence selection algorithm are clearly demonstrated. From the obtained result one can infer that combining all these criteria gives better performance compared to individual criteria and other models individually. The experimental results show that our approach has very promising results and it significantly outperforms many state-of-the-art methods. Those results assert the great ability of LSA algorithm to capture the semantic representation of the document. However, there is a space for further enhancement by improving the weighting schemes of the LSA input matrix. We believe that using word embedding, namely word2vec or GloVe, for this purpose is interesting to try.

---

[3] http://www.berouge.com/Pages/default.aspx. ROUGE-1.5.5 with options: -n 2 -m -u -c 95 -r 1000 -f A -p 0.5 -t 0 -l 100.

[4] http://multiling.iit.demokritos.gr/pages/view/1532/taskmss-single-document-summarization-data-and-information.

[5] ROUGE-1.5.5 with options -n 2 -2 4 -u -x -m.

**Acknowledgements** We are grateful to the support of the National Natural Science Foundation of China (Grant Nos. 61379109, M1321007) and Science and Technology Plan of Hunan Province (Grant Nos. 2014GK2018, 2016JC2011). We would like to thank the anonymous referees for their helpful comments and suggestions.

## A Appendix

**Table 8** Summaries of the document in LDC Arabic Newsware-a dataset using the proposed model(RAWEFPOS4ADJ)

| Original Document | ( احتفالات القدس بمناسبة حلول العام 2000) |
|---|---|
| شهدت القدس بمناسبة حلول العام 2000م، إحتفالات دينية إنتهت في جبل الزيتون وإحتفالات شعبية تخللها إطلاق ألعاب نارية في سماء المدينة وجرت بهدوء.  ولم تتحدث السلطات الإسرائيلية التي كانت تخشى حدوث حالات إنتحار جماعية من قبل مسيحيين ألفيين يعتقدون أنهم سيسرعون بذلك نهاية العالم وعودة المسيح إلى الأرض، عن أي حادث من هذا النوع.  وذكرت الإذاعة الإسرائيلية أنه تم نقل بعض الأشخاص إلى مستشفيات للأمراض النفسية وأن الشرطة منعت عشرات المسيحيين من إشعال شموع قرب حائط المبكى.  وقد شارك حوالي ألفي شخص بينهم كهنة ورجال دين جاؤوا للإحتفال بالذكرى الألفية الثانية لميلاد المسيح، في قداس أقيم في كنيسة في القدس الشرقية على سفح جبل الزيتون مقابل سور البلدة القديمة.  وتلا القداس زياح حتى جبل الزيتون وسط تراتيل باللغة العربية.  وفي منتصف الليل أضاءت ألعاب نارية سماء القدس بينما علت صرخات الفرح بحلول العام 2000. وتحت ضغط الأوساط اليهودية المتشددة للمحافظة على عطلة السبت، ألغي عدد من الإحتفالات التي كان من المقرر تنظيمها في القدس بمناسبة إنتهاء القرن العشرين. ولم تبق السلطات سوى على حفلتين إحداهما في مرقص والثانية في نزل للشبان المسيحيين.  واستقبلت تل ابيب من جهتها العام 2000 بفرح وعلت الألحان الموسيقية وصرخات الفرح في شوارع المدينة التي غصت بالمحتفلين. | |
| Golden Summary | |
| شهدت القدس بمناسبة حلول العام 2000م، إحتفالات دينية إنتهت في جبل الزيتون وإحتفالات شعبية تخللها إطلاق ألعاب نارية في سماء المدينة وجرت بهدوء. وقد شارك حوالي ألفي شخص بينهم كهنة ورجال دين جاؤوا للإحتفال بالذكرى الألفية الثانية لميلاد المسيح، في قداس أقيم في كنيسة في القدس الشرقية على سفح جبل الزيتون مقابل سور البلدة القديمة.  وتلا القداس زياح حتى جبل الزيتون وسط تراتيل باللغة العربية. واستقبلت تل ابيب من جهتها العام 2000 بفرح وعلت الألحان الموسيقية وصرخات الفرح في شوارع المدينة التي غصت بالمحتفلين. | |
| The system Summary | |
| وقد شارك حوالي ألفي شخص بينهم كهنة ورجال دين جاؤوا للإحتفال بالذكرى الألفية الثانية لميلاد المسيح، في قداس أقيم في كنيسة في القدس الشرقية على سفح جبل الزيتون مقابل سور البلدة القديمة.  وفي منتصف الليل أضاءت ألعاب نارية سماء القدس بينما علت صرخات الفرح بحلول العام 2000. وتحت ضغط الأوساط اليهودية المتشددة للمحافظة على عطلة السبت، ألغي عدد من الإحتفالات التي كان من المقرر تنظيمها في القدس بمناسبة إنتهاء القرن العشرين. | |

## References

1. Binwahlan, M.S.; Salim, N.; Suanmali, L.: Fuzzy swarm diversity hybrid model for text summarization. Inf. Process. Manag. **46**(5), 571–588 (2010). https://doi.org/10.1016/j.ipm.2010.03.004
2. Khan, K.; Baharudin, B.B.; Khan, A.: Semantic-based unsupervised hybrid technique for opinion targets extraction from unstructured reviews. Arab. J. Sci. Eng. **39**(5), 3681–3689 (2014). https://doi.org/10.1007/s13369-014-0990-1
3. Imam, I.; Nounou, N.; Hamouda, A.; Khalek, H.A.A.: An ontology-based summarization system for Arabic documents (OSSAD). Int. J. Comput. Appl. **74**(17), 38–43 (2013)
4. Qumsiyeh, R.; Ng, Y.-K.: Enhancing web search by using query-based clusters and multi-document summaries. Knowl. Inf. Syst. **47**(2), 355–380 (2016). https://doi.org/10.1007/s10115-015-0852-5
5. Sarkar, D.: Text Summarization. In: Text Analytics with Python: A Practical Real-World Approach to Gaining Actionable Insights from your Data, pp. 217–263. Apress, Berkeley, CA (2016)
6. Heu, J.-U.; Qasim, I.; Lee, D.-H.: FoDoSu: multi-document summarization exploiting semantic analysis based on social Folksonomy. Inf. Process. Manag. **51**(1), 212–225 (2015). https://doi.org/10.1016/j.ipm.2014.06.003
7. Hammo, B.H.: A hybrid arabic text summarization technique based on text structure and topic identification. Int. J. Comput. Process. Lang. **23**(1), 39–65 (2011)
8. Al Qassem, L.M.; Wang, D.; Al Mahmoud, Z.; Barada, H.; Al-Rubaie, A.; Almoosa, N.I.: Automatic Arabic summarization: a survey of methodologies and systems. Proc. Comput. Sci. **117**, 10–18 (2017). https://doi.org/10.1016/j.procs.2017.10.088
9. Ferreira, R.; de Souza Cabral, L.; Lins, R.D.; Pereirae Silva, G.; Freitas, F.; Cavalcanti, G.D.C.; Lima, R.; Simske, S.J.; Favaro, L.: Assessing sentence scoring techniques for extractive text summarization. Expert Syst. Appl. **40**(14), 5755–5764 (2013). https://doi.org/10.1016/j.eswa.2013.04.023
10. Zhu, J.; Jiang, Y.; Li, B.; Sun, M.: Ontology-based automatic summarization of web document. Int. J. Adv. Comput. Technol. **4**(14), 289–309 (2012). https://doi.org/10.4156/ijact.vol4.issue14.34
11. Jeong, H.; Ko, Y.; Seo, J.: How to improve text summarization and classification by mutual cooperation on an integrated framework. Expert Syst. Appl. **60**, 222–233 (2016). https://doi.org/10.1016/j.eswa.2016.05.001
12. Isonuma, M.; Fujino, T.; Mori, J.; Matsuo, Y.; Sakata, I.: Extractive Summarization Using multi-task learning with document classification. In: Proceedings of the 2017 Conference on Empirical Methods in Natural Language Processing (2017), pp. 2091–2100
13. Triantafillou, E.; Kiros, J.R.; Urtasun, R.; Zemel, R.: Towards generalizable sentence embeddings. In: Proceedings of the 1st Workshop on Representation Learning for NLP, Berlin, Germany, 2016, pp. 239–248
14. Wu, Z.; Lei, L.; Li, G.; Huang, H.; Zheng, C.; Chen, E.; Xu, G.: A topic modeling based approach to novel document automatic summarization. Expert Syst. Appl. **84**, 12–23 (2017). https://doi.org/10.1016/j.eswa.2017.04.054
15. Gong, Y.; Liu, X.: Generic text summarization using relevance measure and latent semantic analysis. In: Proceedings of the 24th Annual International ACM SIGIR Conference on Research and Development in Information Retrieval, pp. 19–25, ACM (2001)
16. Steinberger, J.; Jezek, K.: Using latent semantic analysis in text summarization and summary evaluation. In: Proceedings of the ISIM'04, pp. 93–100 (2004)
17. Murray, G.; Renals, S.; Carletta, J.: Extractive summarization of meeting recordings. In: INTERSPEECH (2005)
18. Ozsoy, M.G.; Alpaslan, F.N.; Cicekli, I.: Text summarization using latent semantic analysis. J. Inf. Sci. **37**(4), 405–417 (2011). https://doi.org/10.1177/0165551511408848
19. Wang, Y.; Ma, J.: A comprehensive method for text summarization based on latent semantic analysis. In: Zhou, G., Li, J., Zhao, D., Feng, Y. (eds.) Natural Language Processing and Chinese Computing: Second CCF Conference, NLPCC 2013, Chongqing, China, November 15–19, 2013, Proceedings, pp. 394–401. Springer, Berlin (2013)
20. Shen, Y.; He, X.; Gao, J.; Deng, L.; Gr, #233, Mesnil, g.: A latent semantic model with convolutional-pooling structure for information retrieval. In: Proceedings of the 23rd ACM International Conference on Conference on Information and Knowledge Management, Shanghai, China, pp. 101–110. ACM, 2661935 (2014)
21. Al-Saleh, A.B.; Menai, M.E.B.: Automatic Arabic text summarization: a survey. Artif. Intell. Rev. **45**(2), 203–234 (2016). https://doi.org/10.1007/s10462-015-9442-x
22. El-Haj, M.; Kruschwitz, U.; Fox, C.: Multi-document Arabic text summarisation. In: 2011 3rd Computer Science and Electronic Engineering Conference (CEEC), pp. 40–44 (2011)
23. Froud, H.; Lachkar, A.; Ouatik, S.A.: Arabic text summarization based on latent semantic analysis to enhance Arabic documents clustering. arXiv preprint arXiv:1302.1612 (2013)
24. Ba-Alwi, F.M.; Gaphari, G.H.; Al-Duqaimi, F.N.: Arabic text summarization using latent semantic analysis. Br. J. Appl. Sci. Technol. **10**(2), 1–14 (2015)
25. Althobaiti, M.; Kruschwitz, U.; Poesio, M.: AraNLP: A Java-Based Library for the Processing of Arabic Text, pp. 4134–4138. University of Essex, Colchester (2013)
26. Farghaly, A.; Shaalan, K.: Arabic natural language processing: challenges and solutions. ACM Trans. Asian Lang. Inf. Process. **8**(4), 1–22 (2009). https://doi.org/10.1145/1644879.1644881
27. Nadera, B.: The Arabic natural language processing: introduction and challenges. Int. J. Engl. Lang. Transl. Stud. **2**(3), 106–112 (2014)
28. El-Khair, I.A.: Effects of stop words elimination for Arabic information retrieval: a comparative study. Int. J. Comput. Inf. Sci. **4**(3), 119–133 (2006)
29. Taghva, K.; Elkhoury, R.; Coombs, J.: Arabic stemming without a root dictionary. In: International Conference on Information Technology: Coding and Computing (ITCC'05)—Volume II, 4–6, vol. 151, pp. 152–157 (2005)
30. Zahedi, M.-H.; Kahani, M.: SREC: discourse-level semantic relation extraction from text. Neural Comput. Appl. **23**, 1573–1582 (2013)
31. Köprü, S.: An efficient part-of-speech tagger for Arabic. In: Gelbukh, A.F. (ed.) Computational Linguistics and Intelligent Text Processing: 12th International Conference, CICLing 2011, Tokyo, Japan, February 20–26, 2011. Proceedings, Part I, pp. 202–213. Springer, Berlin (2011)
32. Manning, C.D.; Raghavan, P.; Schütze, H.: Introduction to Information Retrieval. Cambridge University Press, Cambridge (2008)
33. Lee, J.-H.; Park, S.; Ahn, C.-M.; Kim, D.: Automatic generic document summarization based on non-negative matrix factorization. Inf. Process. Manag. **45**(1), 20–34 (2009). https://doi.org/10.1016/j.ipm.2008.06.002
34. Kalman, D.: A singularly valuable decomposition: the SVD of a matrix. Coll. Math. J. **27**(1), 2–23 (1996)
35. Menéndez, H.D.; Plaza, L.; Camacho, D.: A Genetic Graph-Based Clustering Approach to Biomedical Summarization, pp. 978-1-4503-1850-1. ACM (2013)
36. Jing, H.; Barzilay, R.; McKeown, K.; Elhadad, M.: Summarization evaluation methods: experiments and analysis. In: AAAI Symposium on Intelligent Summarization, pp. 51–59 (1998)

37. Sobh, I.; Darwish, N.; Fayek, M.: Evaluation Approaches for an Arabic Extractive Generic Text Summarization System, pp. 150–155. The Research and Development International Company, RDI, Cairo University, Giza, Egypt. http://www.rdi-eg.com (2013)
38. Lin, C.-Y.: Rouge: a package for automatic evaluation of summaries. In: Text Summarization Branches Out: Proceedings of the ACL-04 Workshop, Barcelona, Spain (2004)
39. El-Haj, M.: Essex Arabic summaries corpus (EASC). In: Text Analysis Conference (TAC) 2011, vol. 2016, vol. 10/03/2015. Lancaster University (2011)
40. El-Haj, M.; Kruschwitz, U.; Fox, C.: Using mechanical turk to create a corpus of Arabic summaries. In: Proceedings of the International Conference on Language Resources and Evaluation. European Language Resources Association (2010)
41. Lin, C.-Y.; Hovy, E.: Automatic evaluation of summaries using N-gram co-occurrence statistics. In: Proceedings of the 2003 Conference of the North American Chapter of the Association for Computational Linguistics on Human Language Technology—Volume 1, Edmonton, Canada 2003, pp. 71–78. Association for Computational Linguistics, 1073465
42. Giannakopoulos, G.; Kubina, J.; Conroy, J.; Steinberger, J.; Favre, B.; Kabadjov, M.; Kruschwitz, U.; Poesio, M.: Multiling 2015: multilingual summarization of single and multi-documents, on-line fora, and call-center conversations. In: Proceedings of SIGDIAL, Prague, pp. 270–274 (2015)